\documentclass[runningheads]{llncs}
\usepackage{graphicx}
\usepackage{tabularx}
\usepackage{booktabs}
\usepackage{bm}
\usepackage{amsmath}
\usepackage{multirow}
\usepackage{makecell}
\usepackage{subfigure}
\DeclareMathOperator*{\argmax}{arg\,max}

\begin{document}
\title{Multi-Task Learning based Online Dialogic Instruction Detection with Pre-trained Language Models}
\titlerunning{Multi-Task Learning based Online Dialogic Instruction Detection}
%

\author{Yang Hao\inst{1}, Hang Li \inst{1}, Wenbiao Ding \inst{1}, Zhongqin Wu \inst{1}, Jiliang Tang \inst{2}, Rose Luckin \inst{3}, Zitao Liu \inst{1}\thanks{Corresponding Author: Zitao Liu}}

\authorrunning{Y. Hao et al.}
%

\institute{TAL Education Group, Beijing, China \email{\{haoyang2,lihang4,dingwenbiao,wuzhongqin,liuzitao\}@tal.com} \and
Data Science and Engineering Lab, Michigan State University, USA
\email{tangjili@msu.edu} \and
UCL Knowledge Lab, London, UK \\
\email{r.luckin@ucl.ac.uk} 
}

\maketitle              

\begin{abstract}
\label{sec:abstract}
In this work, we study computational approaches to detect online dialogic instructions, which are widely used to help students understand learning materials, and build effective study habits. This task is rather challenging due to the widely-varying quality and pedagogical styles of dialogic instructions. To address these challenges, we utilize pre-trained language models, and propose a multi-task paradigm which enhances the ability to distinguish instances of different classes by enlarging the margin between categories via contrastive loss. Furthermore, we design a strategy to fully exploit the misclassified examples during the training stage. Extensive experiments on a real-world online educational data set demonstrate that our approach achieves superior performance compared to representative baselines. To encourage reproducible results, we make our implementation online available at \url{https://github.com/AIED2021/multitask-dialogic-instruction}.
\end{abstract}

\section{Introduction}
\label{sec:intro}
Teaching online classes is a very challenging task for classroom instructors trained to work offline \cite{huang2020neural,xu2020automatic}. When sitting in front of a camera or a laptop, traditional classroom instructors lack effective pedagogical instructions to ensure the overall quality of their online classes \cite{chen2019multimodal,li2020multimodal}. In this paper, we develop a set of dialogic instructions for online classes aiming to encourage talks and discourses between teachers and students, in addition to teacher-presentation \cite{haghverdi2010note,lee2008effects,henderlong2002effects,dweck2007boosting}. Furthermore, we study computational approaches to automatically detect these dialogic instructions from online class videos, which provides timely feedback to teachers and help them improve their online teaching skills. 

However, automatic dialogic instruction detection poses numerous challenges in real-life teaching scenarios. First, online teaching is not a standardized procedure. Even for the same learning content, different instructors may teach it in various ways according to their own pedagogical styles. Furthermore, the quality of dialogic instructions varies a lot from junior to senior instructors. The second challenge is that the model has to be robust enough to errors from automatic speech recognition (ASR) transcriptions. Publicly available ASR services may yield very high transcription errors, which lead to inferior performance in the noisy and dynamic classroom environments \cite{blanchard2015study}.

To address the above challenges, in this study, we propose an end-to-end multi-task framework for automatic dialogic instruction detection from online videos. Specifically, we (1) propose a contrastive loss based multi-task framework to distinguish instances by enlarging the distances between instances of different categories; (2) utilize the pre-trained neural language model to robustly handle errors from ASR transcriptions without the need for manual annotation efforts; and (3) propose a strategy to select and exploit hard instances in the training process to achieve higher performance. 

\section{The Dialogic Instruction Detection Framework}
\label{sec:method}
In this work, we aim to capture the following eight types of well-studied dialogic instructions that (1) motivate students and make them feel easy about the class: \textit{greeting} \cite{goodenow1993psychological,osterman2010teacher} and \textit{commending} \cite{henderlong2002effects,dweck2007boosting}, (2) help students understand learning materials and retain them: \textit{guidance} \cite{yelland2007rethinking}, \textit{example-giving} \cite{shafto2014rational}, \textit{repeating} \cite{anthony2015supporting}, and \textit{reviewing} \cite{an2004capturing}, and (3) build effective learning habits:  \textit{note-taking} \cite{haghverdi2010note,lee2008effects} and \textit{summarization} \cite{rinehart1986some}.
	
Our multi-task dialogic instruction detection framework has three key components: (1) a pre-trained language model, which serves as the base model in the classification task; (2) a multi-task learning module, which distinguishes effective instructions from similar but ineffective ones by pushing instances from different categories apart; and (3) a hard example mining strategy, which establishes a hard example set to select instances when constructing input pairs. 

\noindent \textbf{Pre-trained Language Model} To extract contextual information, in this study we utilize the Transformer-based pre-trained language model as our base model in our detection framework. To perform the instruction detection task on a sentence, similar to \cite{devlin2018bert,liu2019roberta}, we first add a special token $[CLS]$ in front of the sentence. After that, sentences are fed into multiple Transformer encoders sequentially. Finally the hidden state of the special token $[CLS]$ from the last layer of Transformer encoders is obtained as the representation of the sentence.

\noindent \textbf{Multi-task Learning Module} The multi-task learning framework consists of two sub-tasks: (1) a multi-class classification task to decide which category a dialogic instruction belongs to, where the cross-entropy loss is used; and (2) an additional task with an objective to enlarge the distances between pairs of instructions from different categories by using contrastive loss. The total loss is a combination of the two parts above defined as follows:


\begin{equation} \nonumber
L =  \gamma \cdot \underbrace{\sum_{i=1}^{b} \sum_{c \in \mathbf{C}} -y_i^c \cdot \log(\hat{y}^c_i)}_{\text{\normalfont cross-entropy loss}}+ (1 - \gamma) \cdot \underbrace{ \sum_{i=1}^{b} \big( \max \big\{ 0, M - \| \mathcal{F}_\Theta(\mathbf{x}_i) - \mathcal{F}_\Theta(\mathbf{x}_j^{\tilde{g}}) \|_2 \big\} \big)^2 }_{\text{\normalfont contrastive loss}} 
\end{equation}

\noindent where $\mathbf{x}_i$ denotes the raw feature of the $i$th instance and $y_i^c$ represents the indicator variable that is equal to 1 if and only if the $i$th instance belongs to the ground truth category $g$. $\hat{y}^c_i$ is the predicted probability that the $i$th instance belongs to category $c$ and $b$ is the batch size. $\mathcal{F}_\Theta(\cdot)$ denotes the pre-trained language model, which extracts representation of an input instance.  $\gamma$ and $M$ are hyper-parameters. $\mathbf{x}_j^{\tilde{g}}$ denotes an arbitrary instance (indexed by $j$) that comes from a different category of $\mathbf{x}_i$, and $\tilde{g} = \mathbf{C} \backslash \{g\}$.

\noindent \textbf{Hard Example Mining Strategy} Instances easily classified correctly by the model contribute little to the contrastive loss \cite{schroff2015facenet,wang2020representation}. That is to say, a randomly selected instance $\mathbf{x}_j^{\tilde{g}}$ probably has been far away from an instance $\mathbf{x}_i$ after epochs of training. Therefore, instead of generating pairs by random sampling, we focus on hard examples, i.e., instances that are misclassified into a wrong category. Hence, the hard example set $\mathbf{H}$ is discovered by: $\mathbf{H} = \{ \textbf{x}_j | \argmax{y_j}  \neq \argmax{\hat{y}_j}, j = 1, \cdots, b \}$. Pairs of training inputs are selected by first randomly choosing an instance $\mathbf{x}_i$ from the entire training set $\mathbf{X}$, and then randomly choosing $\mathbf{x}_j^{\tilde{g}}$ from the hard example set $\mathbf{H}$.

\vspace{-0.2cm}
\section{Experiments}
\label{sec:experiments}
\vspace{-0.2cm}
We collected online-class video recordings from a third-party educational platform. Similar to \cite{xu2020automatic,huang2020neural}, audio tracks are extracted from video recordings and then cut into utterances by a self-trained VAD model \cite{tashev2016dnn}. After that, utterances are transcribed into text using a self-trained ASR model \cite{zhang2018deep} with a character error rate of 11.36\% in classroom scenarios. The training and validation sets contains 16,174 and 4,088 instances respectively. Performance on each category (except \textit{others}) is separately evaluated on a binary test set containing 2000 positive instances that belong to this category, and 2000 negative ones from the other categories (other seven categories of instructions, or \textit{others}). We select a series of widely-used baselines, including BiLSTM \cite{graves2013speech}, TextRCNN \cite{lai2015recurrent}, and pre-trained language models: BERT \cite{devlin2018bert}, ELECTRA \cite{clark2020electra}, NEZHA \cite{wei2019nezha}, RoBERTa \cite{liu2019roberta}, and XLNet \cite{yang2019xlnet}. Moreover, we compare different strategies of negative example selection in our multi-task framework: (1) random selection from all the instances of other categories, i.e., \textit{M-RoBERTa-All}; and (2) hard example mining, i.e., \textit{M-RoBERTa-Hard}. 

\vspace{-0.2cm}
\subsection{Results Discussion}
\label{sec:performance}

\begin{table}[!t]
	\centering
	\caption{\label{tab:performance} Performance of different pre-trained language models.}  \vspace{-0.3cm}
	\resizebox{0.77\textwidth}{!}{
		\begin{tabular}{clccclcc}
			\toprule
			\multicolumn{1}{c}{Instruction} & Model          & Accuracy       & F1             & \multicolumn{1}{c}{Instruction} & Model          & Accuracy       & F1             \\
			\midrule
			\multirow{7}{*}{macro-average}  & BiLSTM         & 0.781          & 0.783          & \multirow{7}{*}{micro-average}  & BiLSTM         & 0.781          & 0.791          \\
			& TextRCNN       & 0.785          & 0.788          &                                 & TextRCNN       & 0.785          & 0.789          \\
			& BERT           & 0.781          & 0.787          &                                 & BERT           & 0.781          & 0.778          \\
			& ELECTRA        & 0.791          & 0.790           &                                 & ELECTRA        & 0.791          & 0.794          \\
			& NEZHA          & 0.797          & 0.803          &                                 & NEZHA          & 0.797          & \textbf{0.797}          \\
			& XLNet          & 0.770           & 0.775          &                                 & XLNet          & 0.770           & 0.764          \\
			& RoBERTa        & \textbf{0.799}          & \textbf{0.812}          &                                 & RoBERTa        & \textbf{0.799}          & 0.795          \\
			\bottomrule
	\end{tabular}}
\end{table}

\begin{table}[!t]
	\centering
	\caption{\label{tab:multitask_performance} Performance of the proposed method and its variants.}  \vspace{-0.3cm}
	\resizebox{0.77\textwidth}{!}{
		\begin{tabular}{clccclcc}
			\toprule
			\multicolumn{1}{c}{Instruction} & Model          & Accuracy       & F1             & \multicolumn{1}{c}{Instruction} & Model          & Accuracy       & F1             \\
			\midrule
			\multirow{3}{*}{commending}     & RoBERTa        & 0.828          & 0.831          &       \multirow{3}{*}{guidance}                          & RoBERTa        & 0.809          & 0.829          \\
			& M-RoBERTa-All      & 0.831          & 0.844          &                                 & M-RoBERTa-All      & 0.847          & 0.850           \\
			& M-RoBERTa-Hard & \textbf{0.842}          & \textbf{0.855} &                                 & M-RoBERTa-Hard & \textbf{0.868} & \textbf{0.872} \\
			\midrule
			\multirow{3}{*}{summarization}	& RoBERTa        & 0.803          & 0.829          &                   \multirow{3}{*}{greeting}               & RoBERTa        & 0.788          & 0.803          \\
			& M-RoBERTa-All      & 0.862          & 0.875          &                                 & M-RoBERTa-All      & 0.791          & 0.810           \\
			& M-RoBERTa-Hard & \textbf{0.876} & \textbf{0.886} &                                 & M-RoBERTa-Hard & \textbf{0.802}          & \textbf{0.830}  \\
			\midrule
			\multirow{3}{*}{note-taking} & RoBERTa        & 0.814          & 0.830           &              \multirow{3}{*}{repeating}                      & RoBERTa        & 0.690           & 0.725          \\
			& M-RoBERTa-All      & 0.735          & 0.771          &                                 & M-RoBERTa-All      & 0.749          & 0.774          \\
			& M-RoBERTa-Hard & \textbf{0.886} & \textbf{0.889} &                                 & M-RoBERTa-Hard & \textbf{0.750}           & \textbf{0.776}          \\
			\midrule
			\multirow{3}{*}{reviewing}  	& RoBERTa        & 0.796          & 0.787          &                  \multirow{3}{*}{example-giving}                & RoBERTa        & 0.868          & 0.859          \\
			& M-RoBERTa-All      & \textbf{0.824}          & \textbf{0.811}          &                                 & M-RoBERTa-All      & 0.861          & 0.854          \\
			& M-RoBERTa-Hard & 0.822          & \textbf{0.811}          &                                 & M-RoBERTa-Hard & \textbf{0.929} & \textbf{0.893}          \\
			\midrule
			\multirow{3}{*}{macro-average} & RoBERTa        & 0.799          & 0.812          &                   \multirow{3}{*}{micro-average}              & RoBERTa        & 0.799          & 0.795          \\
			& M-RoBERTa-All      & 0.812          & 0.824          &                                 & M-RoBERTa-All      & 0.812          & 0.804          \\
			& M-RoBERTa-Hard & \textbf{0.847} & \textbf{0.852} &                                 & M-RoBERTa-Hard & \textbf{0.847} & \textbf{0.823} \\
			\bottomrule
	\end{tabular}}
	\vspace{-0.5cm}
\end{table}

From Table \ref{tab:performance}, we can find that pre-trained language models such as ELECTRA, NEZHA, and RoBERTa achieve better performance than classic approaches, i.e., BiLSTM and TextRCNN, which indicates their stronger capacity to model dialogic instructions by utilizing contextual information. ELECTRA, RoBERTa, and NEZHA have a better overall performance than BERT, which is not surprising since they are pre-trained with improved training objectives and larger corpus. 

We demonstrate the effectiveness of our multi-task framework by comparing with standard RoBERTa model (in Table \ref{tab:performance}). Table \ref{tab:multitask_performance} shows that: (1) by adding a contrastive loss to enlarge the margin between different categories, \textit{M-RoBERTa-All} outperforms the original RoBERTa model in 6 out of 8 types of dialogic instructions and the overall performance; and (2) by fully utilizing instances misclassified by the model, \textit{M-RoBERTa-Hard} outperforms \textit{M-RoBERTa-All} and achieves the best prediction performance compared with other methods in terms of accuracy, macro- and micro-F1 scores.

\vspace{-0.1cm}
\section{Conclusion}
\label{sec:conclusion}
We present a multi-task dialogic instruction detection framework using pre-trained language models. Furthermore, we design a strategy to select hard instances and exploit them during training. Experiments conducted on a real-world data set show that our framework outperforms both classic methods and pre-trained language models fine-tuned solely with the classification objective.

\vspace{-0.1cm}
\section*{Acknowledgment}

This work was supported in part by National Key R\&D Program of China, under Grant No. 2020AAA0104500 and in part by Beijing Nova Program (Z201100006820068) from Beijing Municipal Science \& Technology Commission.

\bibliographystyle{splncs04}
\bibliography{aied2021.bib}

\end{document}